\documentclass[11pt]{article}

\usepackage[preprint]{acl}

\usepackage{times}
\usepackage{latexsym}

\usepackage[T1]{fontenc}

\usepackage[utf8]{inputenc}

\usepackage{microtype}

\usepackage{inconsolata}

\usepackage{graphicx}

\usepackage{todonotes}
\usepackage{booktabs}
\usepackage{array}
\usepackage{tabularx}
\usepackage{multirow}
\usepackage{makecell}
\usepackage{threeparttable}
\usepackage{amsmath}
\usepackage{amssymb}
\usepackage{enumitem}
\usepackage{placeins}
\usepackage{CJKutf8}
\usepackage{tcolorbox}
\tcbuselibrary{breakable}

\newenvironment{promptbox}[1]{%
  \begin{tcolorbox}[breakable,colback=gray!5,colframe=black!45,title={#1},fonttitle=\bfseries,boxsep=1mm,left=1mm,right=1mm,top=1mm,bottom=1mm]%
  \footnotesize
  \begin{CJK*}{UTF8}{gbsn}%
}{%
  \end{CJK*}%
  \end{tcolorbox}%
}

\title{Personalized Turn-Level User Conversation Satisfaction Benchmark}

\author{
  \textbf{Zhefan Wang\textsuperscript{1}},
  \textbf{Zhiqiang Guo\textsuperscript{1}},
  \textbf{Weizhi Ma\textsuperscript{2}}\footnotemark[1],
  \textbf{Min Zhang\textsuperscript{1}}\footnotemark[1],
  \textbf{Quanjia Yan\textsuperscript{3}},
  \textbf{Hengliang Luo\textsuperscript{3}} \\
  \\
  \textsuperscript{1}Department of Computer Science and Technology, Tsinghua University, Beijing, China. \\
  \textsuperscript{2}Institute for AI Industry Research, Tsinghua University, Beijing, China.
  \textsuperscript{3}Meituan. \\
  \small{
    \texttt{thuwzf2000@gmail.com, mawz@tsinghua.edu.cn, z-m@tsinghua.edu.cn}
  }
}

\begin{document}
\maketitle
\footnotetext[1]{Corresponding author.}
\begin{abstract}
User satisfaction with AI assistants is highly personalized: the same response may satisfy one user but disappoint another depending on what each user expects and what they have asked for before.
Existing automatic evaluation methods mostly measure generic response quality, making it difficult to judge whether a response satisfies a user at a specific turn.
We study this problem as personalized turn-level user conversation satisfaction evaluation.
We build a conversation satisfaction evaluator that combines compact user memories with target-turn context to produce satisfaction scores and dissatisfaction-oriented rationales.
Meta-evaluation against human satisfaction annotations shows that personalized memory and post-hoc score calibration improve ordinal agreement and dissatisfied-turn detection over supervised, retrieval-based, and generic LLM-as-a-judge baselines.
We further introduce PersTurnBench, a personalized turn-level user conversation satisfaction benchmark that uses the verified evaluator to assess generation models via replay.
By holding the replay state fixed, PersTurnBench enables controlled comparison of generic generation models and memory-augmented personalized systems without new human labels for every candidate model.
The evaluator and benchmark let researchers compare candidate generation models on personalized satisfaction without collecting new user feedback for every model.
\end{abstract}

\section{Introduction}

Large language model assistants are increasingly used in task-oriented and personalized conversations, such as planning trips, organizing learning plans, and supporting everyday decisions~\citep{ouyang2022training,stiennon2020learning,salemi2024lamp,zhang2018personalizing}.
Evaluating such assistants is not only about judging whether a response is fluent, helpful, correct, or generally useful.
In practical assistant settings, user satisfaction also reflects preferences, expectations, constraint sensitivity, and signals from past interactions.
For example, a concise recommendation may satisfy a user who prefers direct answers, while disappointing another user who expects detailed steps, concrete resources, or explicit handling of constraints.
This makes personalized turn-level satisfaction an important evaluation target.

Existing automatic evaluation work has largely focused on generic generation quality or task-oriented success, with comparison-based preference judgments as a recent addition.
Traditional generation metrics and recent LLM-as-a-judge methods can compare responses with respect to fixed quality criteria~\citep{papineni2002bleu,lin2004rouge,zhang2019bertscore,zheng2023judging,liu2023g}.
Task-oriented conversation benchmarks commonly evaluate state tracking, response generation, or task completion~\citep{budzianowski2018multiwoz,rastogi2020towards}.
These methods are valuable, but they do not directly answer whether a specific assistant turn satisfies a specific user.
The turn-level distinction matters because user intents and constraints can shift within a conversation, and a few key turns may dominate the final experience.
The personalized distinction matters because users differ in communication style, detail preference, and sensitivity to task constraints, so the same response can be acceptable for one user and unsatisfactory for another.
Prior work on task-oriented conversation satisfaction has shown that satisfaction is meaningful but challenging when evaluation is local and context-sensitive~\citep{sun2021simulating,li2021deus,deng2022user}.
These limitations motivate an evaluator that can use user evidence from previous interactions and judge satisfaction at the assistant-turn level.

In this work, we study personalized turn-level user conversation satisfaction evaluation.
Given a target conversation state, a candidate assistant turn, and user evidence from previous interactions, the goal is to estimate the satisfaction score that the turn would receive from that user.
We build a conversation satisfaction evaluator that converts past ratings and interaction histories into user memories for future turn-level assessment.
The evaluator summarizes user history, judges target turns with personalized evidence, and calibrates scores to reduce mismatch between raw LLM scores and user-specific rating scales.
We verify the evaluator through meta-evaluation against human satisfaction annotations and compare it with supervised, retrieval-based, and generic LLM-as-a-judge baselines.
After verification, we use the evaluator to construct PersTurnBench, a personalized turn-level user conversation satisfaction benchmark for replay-based model comparison.
Unlike benchmarks for overall ability, task completion, or population-averaged preference, PersTurnBench asks how satisfying candidate responses are expected to be under fixed personalized conversation states.
It provides a lower-cost and reproducible screening layer before recruiting users for new interactions and turn-level feedback, while not serving as a replacement for human evaluation.

In summary, the contributions of our paper are as follows\footnote{Code and data are available at \url{https://github.com/wzf2000/PersTurnBench}.}:
\begin{itemize}[leftmargin=*, parsep=2pt]
    \item We formulate personalized turn-level user conversation satisfaction evaluation and build a conversation satisfaction evaluator that uses user history, conversation context, and assistant responses to estimate satisfaction scores.
    \item Meta-evaluation against human satisfaction annotations shows how our designs affect ordinal agreement and dissatisfied-turn detection.
    \item We introduce PersTurnBench, a turn-level conversation benchmark for comparing candidate generation models under fixed personalized conversation states.
\end{itemize}

\section{Related Work}


\begin{table}[t]
\centering
\resizebox{\columnwidth}{!}{
\begin{threeparttable}
\begin{tabular}{@{}lcccc@{}}
\toprule
Method & Pers. & Turn & \makecell{LLM\\Data} & Direct \\
\midrule
PARADISE~\citep{walker1997paradise} & $\times$ & $\times$ & $\times$ & $\checkmark$ \\
SessionPred~\citep{yao2020session} & $\times$ & $\times$ & $\times$ & $\times$ \\
DEUS~\citep{li2021deus} & $\times$ & $\checkmark$ & $\times$ & $\times$ \\
USDA~\citep{deng2022user} & $\times$ & $\checkmark$ & $\times$ & $\times$ \\
G-Eval~\citep{liu2023g} & $\times$ & $\times$ & $\checkmark$ & $\times$ \\
MT-Bench~\citep{zheng2023judging} & $\times$ & $\times$ & $\checkmark$ & $\times$ \\
P-Judge~\citep{dong2024can} & $\checkmark$ & $\times$ & $\checkmark$ & $\times$ \\
P-ToolEval~\citep{hao2025evaluating} & $\checkmark$ & $\times$ & $\checkmark$ & $\times$ \\
RecUserSim~\citep{chen2025recusersim} & $\checkmark$ & $\times$ & $\checkmark$ & $\times$ \\
\midrule
\textbf{PersTurnBench} & $\checkmark$ & $\checkmark$ & $\checkmark$ & $\checkmark$ \\
\bottomrule
\end{tabular}
\end{threeparttable}
}
\caption{Comparison with representative methods and benchmarks.
Pers. means personalized user evidence, Turn means turn-level satisfaction judging, LLM Data means LLM-era conversation data, and Direct means direct feedback from the original users.}
\vspace{-1em}
\label{tab:related-comparison}
\end{table}

\subsection{Conversation Satisfaction Estimation}

User satisfaction has long been used to evaluate task-oriented conversation systems, from early spoken-conversation frameworks~\citep{walker1997paradise} to recent neural satisfaction models~\citep{yao2020session,bodigutla2020joint,feng2023schema,siro2022understanding}.
Conversation- or session-level estimation captures overall experience but cannot identify which assistant turn causes satisfaction or dissatisfaction.
Turn-level and context-sensitive signals, such as USS~\citep{sun2021simulating}, DEUS~\citep{li2021deus}, and USDA~\citep{deng2022user}, make local failures more visible.
Some settings further use user simulation or interaction rollouts for interactive evaluation~\citep{sun2021simulating,afzali2023usersimcrs,fang2024multi,zhu2024reliable,chen2025recusersim}.

However, conversation-level and rollout-level outcomes can dilute the few key turns that determine user satisfaction and cannot isolate whether a specific assistant turn satisfies a specific user.
Our work instead uses direct satisfaction feedback from the original users and formulates satisfaction evaluation as a personalized turn-level problem.

\subsection{Automatic Conversation Evaluation}

Automatic evaluation is widely used because human evaluation is expensive and hard to repeat for every new candidate model.
Non-LLM metrics such as BLEU, ROUGE, and BERTScore provide scalable text evaluation~\citep{papineni2002bleu,lin2004rouge,zhang2019bertscore}, while conversation-specific metrics estimate relevance, fluency, coherence, or fine-grained quality beyond surface overlap~\citep{tao2018ruber,mehri2020usr,mehri2020unsupervised}.
However, these metrics remain limited proxies when assistant responses must satisfy underspecified user expectations.
Recent LLM-as-a-judge methods, including G-Eval~\citep{liu2023g}, MT-Bench and Chatbot Arena~\citep{zheng2023judging}, and Prometheus~\citep{kim2024prometheus}, offer more flexible rubric- or comparison-based evaluation.
They are not automatically reliable, since prior studies report a range of judge-specific biases such as criterion mismatch and position bias~\citep{li2025exploring,shi2025judging}.

Nevertheless, most automatic evaluation methods still focus on generic response quality or population-level preference under fixed criteria.
Personalized evaluation is a newer direction: an evaluator should reflect what a specific user prefers and expects rather than a population average.
We follow the scalable automatic evaluation paradigm, but adapt it into a conversation satisfaction evaluator that estimates personalized turn-level satisfaction rather than generic quality.

\subsection{Personalized Evaluation}

Personalization research conditions models on profiles, histories, preferences, or memory to produce more personalized outputs~\citep{zhang2018personalizing,salemi2024lamp,jang2023personalized,jiang2025know,jiang2025personamem}.
Related work also studies profile generation and retrieval over personal contexts~\citep{salemi2024lamp,zhang2024guided,jiang2025know}.
These studies show the value of user context, but they target generation or profiling rather than evaluation.
Personalized evaluation has begun to appear in LLM-based persona judging and tool-augmented LLM benchmarks~\citep{dong2024can,hao2025evaluating}, while user simulation evaluates conversational recommendation and assistant systems through synthetic interaction~\citep{afzali2023usersimcrs,fang2024multi,zhu2024reliable,chen2025recusersim}.
Such methods can test interactive behavior, but simulated or persona-based feedback may not match direct satisfaction feedback from the original users who participated in the conversations.

Our setting instead uses historical satisfaction feedback from the original users as personalized evidence and verifies the evaluator against held-out human satisfaction annotations.
PersTurnBench then uses a replay protocol: candidate models generate responses for fixed conversation states, and a frozen conversation satisfaction evaluator scores each response under the same personalized context.
This preserves the scalability of automatic evaluation while keeping the comparison controlled at the user-turn level and grounded in user-specific satisfaction rather than simulated preferences.

\begin{figure*}
  \centering
  \includegraphics[width=\linewidth]{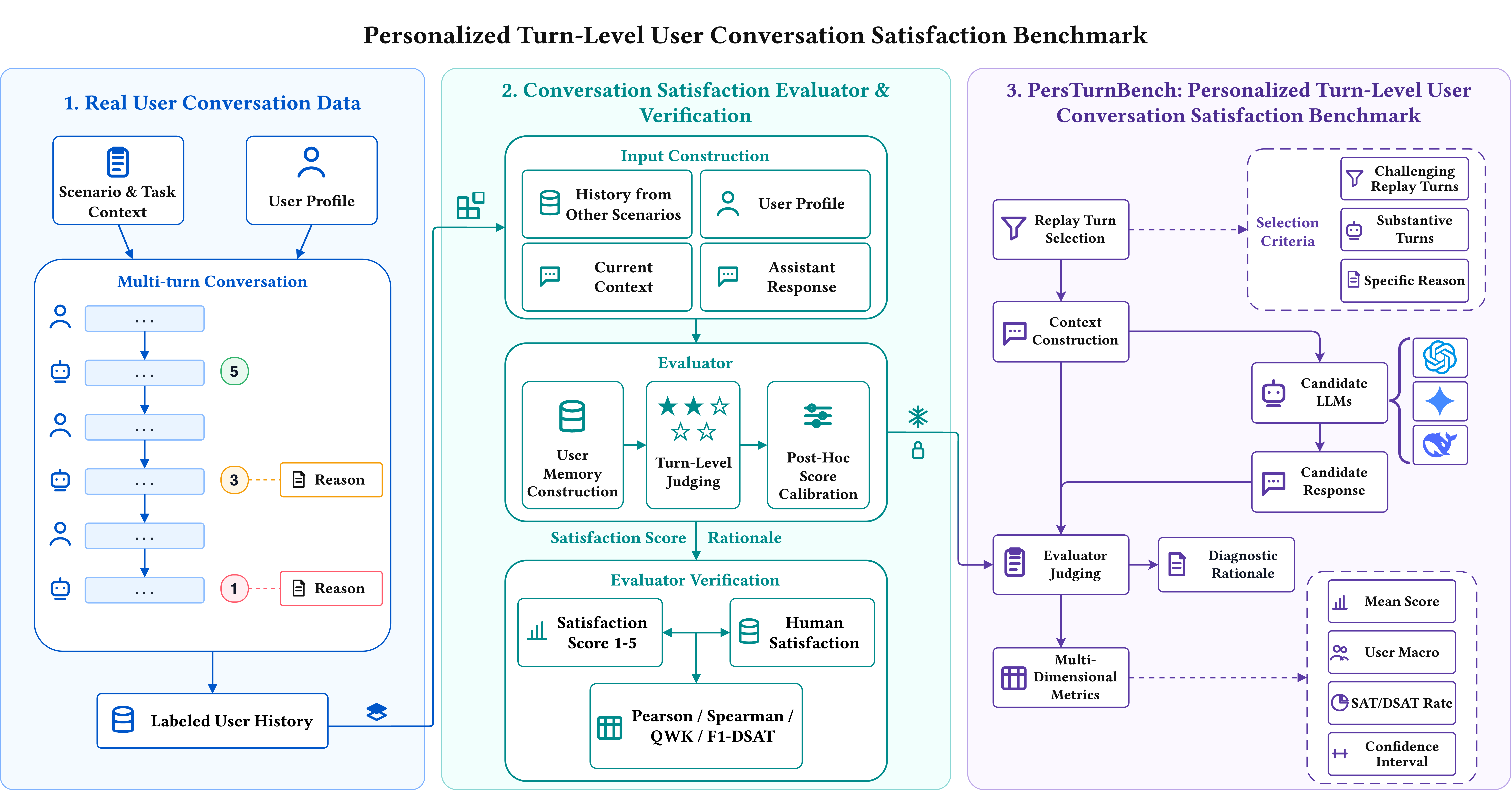} 
  \caption{Overview of our personalized turn-level user conversation satisfaction evaluation framework.
  Real user conversation data provide scenario- and task-conditioned multi-turn conversations, user profiles, turn-level satisfaction scores, and dissatisfaction reasons.
  The conversation satisfaction evaluator constructs its input from the user's source-scenario history, user profile, current conversation context, and the candidate assistant response, then builds structured user memory, judges turn-level satisfaction, and applies post-hoc score calibration to produce a calibrated satisfaction score and a diagnostic rationale.
  We verify this evaluator against human satisfaction annotations, and then use the verified evaluator in PersTurnBench to score candidate LLM responses under controlled replay states and compute personalized benchmark metrics across users, tasks and turns.}
  \label{fig:framework}
  \vspace{-1em}
\end{figure*}

\section{Conversation Satisfaction Evaluator}
\label{sec:evaluator}

A verified conversation satisfaction evaluator allows us to evaluate candidate generation models without collecting new human satisfaction labels for every candidate response.
Figure~\ref{fig:framework} gives the overall workflow, from real user conversations and evaluator verification to replay-based benchmark evaluation with a frozen evaluator.
The evaluator is first verified against human satisfaction annotations in Section~\ref{sec:evaluator-verification}, and then reused as the frozen judge in PersTurnBench in Section~\ref{sec:persturnbench}.
This section defines the personalized turn-level evaluation problem and instantiates it with the memory-based evaluator used in our experiments.

\subsection{Problem Formulation and Notation}
\label{sec:task-formulation}

We use the notation summarized in Appendix~\ref{sec:appendix-notation}.
At a high level, each user $u$ has a profile $p_u$ and multiple task-conditioned conversation sessions $D_{u,s,r}$ under scenario $s$.
When evaluating a target scenario, the evaluator can use the user's source-scenario history $\mathcal{H}_{u}^{-s}$ but cannot access target-scenario gold satisfaction labels.

The basic unit is a target assistant-turn example $\ell=(u,s,r,i)$.
It consists of prior conversation context $C_{\ell}$, current user request $q_{\ell}$, assistant response $a_{\ell}$, task context $\tau_{s_{\ell},r_{\ell}}$, user profile $p_{u_{\ell}}$, and human satisfaction score $y_{\ell}\in\{1,2,3,4,5\}$.
We define the binary satisfaction label at the boundary between scores 3 and 4:
\begin{equation}
z_{\ell}=\mathbb{I}[y_{\ell}\geq 4],
\end{equation}
where $z_{\ell}=1$ denotes a satisfied turn and $z_{\ell}=0$ denotes a dissatisfied turn.

The personalized turn-level satisfaction evaluator maps these inputs to a raw score:
\begin{equation}
\hat{y}_{\ell}
=
\mathcal{E}(C_{\ell},q_{\ell},a_{\ell},p_{u_{\ell}},\tau_{s_{\ell},r_{\ell}},\mathcal{H}_{u_{\ell}}^{-s_{\ell}}),
\end{equation}
We evaluate both the full 1--5 score and the induced satisfied/dissatisfied label $\hat{z}_{\ell}=\mathbb{I}[\hat{y}_{\ell}\geq 4]$.
We use $\hat{y}_{\ell}$ for the raw evaluator output and $\tilde{y}_{\ell}$ for a post-hoc calibrated score.

\subsection{User Memory Construction}

We instantiate $\mathcal{E}$ as a training-free memory-based evaluator.
For a user $u$ and target scenario $s$, the memory builder receives the user profile $p_u$ and the source history $\mathcal{H}_{u}^{-s}$ defined in Section~\ref{sec:data-protocol}.
Each source session contains the task context, the conversation trajectory, turn-level satisfaction scores, and optional dissatisfaction reasons for low-scored assistant turns.
The builder summarizes these labeled histories into a structured memory:
\begin{equation}
  m_u^{-s}=g_{\theta}(\mathcal{H}_{u}^{-s},p_u).
\end{equation}
The memory schema contains three types of information.
First, it stores score-scale statistics, including the historical mean satisfaction score and the 1--5 score distribution.
Second, it stores comparative threshold descriptions, including the user's minimum satisfaction threshold between scores 3 and 4 and excellence threshold between scores 4 and 5.
Third, it stores user-specific requirements, preferred response format, and task-specific observations from prior interactions.

We deliberately make the memory comparative rather than descriptive.
During memory construction, historical assistant turns are grouped by satisfaction score, and the builder is asked to compare adjacent score levels.
The comparison between scores 3 and 4 captures what makes a response cross the minimum satisfaction threshold, while the comparison between scores 4 and 5 captures what turns an acceptable response into an excellent one for this user.
This design turns historical labels into an executable scoring rubric instead of a generic user summary.


\subsection{Turn-Level Judging}

For a target assistant-turn example $\ell$, the judge receives the user memory $m_{u_{\ell}}^{-s_{\ell}}$, the user profile $p_{u_{\ell}}$, the task context $\tau_{s_{\ell},r_{\ell}}$, the prior conversation context $C_{\ell}$, the current user request $q_{\ell}$, and the assistant response $a_{\ell}$.
Conditioning on these inputs, the judge predicts a structured output:
\begin{equation}
  \hat{y}_{\ell},d_{\ell}
  =
  \mathcal{J}_{\phi}(C_{\ell},q_{\ell},a_{\ell},p_{u_{\ell}},\tau_{s_{\ell},r_{\ell}},m_{u_{\ell}}^{-s_{\ell}}),
\end{equation}
where $\hat{y}_{\ell}\in\{1,2,3,4,5\}$ is the predicted score and $d_{\ell}$ is a natural-language diagnostic rationale.
The judge follows a rubric-style decision order.
It first checks whether the response crosses the user's minimum satisfaction threshold between scores 3 and 4.
If the response passes this threshold, it then checks whether the response reaches the user's excellence threshold between scores 4 and 5.
If the response does not pass the minimum satisfaction threshold, it assigns a score from 1 to 3 according to defect severity and the user's specific requirements.

For diagnostic consistency, the rationale is conditioned on the predicted satisfied or dissatisfied class.
For satisfied turns, the evaluator provides a short positive diagnosis.
For dissatisfied turns, it describes the main reason for dissatisfaction using the predefined reason taxonomy.
This diagnostic output is auxiliary and is not used as a primary metric in the main experiments.
Appendix~\ref{sec:appendix-rationale-case-study} provides qualitative examples for rationale case study.

In the main experiments, the memory is built once for each user--target-scenario block.
It is kept fixed when evaluating target turns and is not updated with target-scenario predictions, which avoids using potentially noisy predicted labels as additional evidence.


\subsection{Post-Hoc Score Calibration}

The raw judge output is conditioned on the user's memory, but LLM judges can still be miscalibrated on the absolute 1--5 scale.
We therefore consider post-hoc calibration functions that map raw scores to a user-specific score scale:
\begin{equation}
  \tilde{y}_{\ell}=\kappa(\hat{y}_{\ell};\mathcal{H}_{u_{\ell}}^{-s_{\ell}},\mathcal{I}_{u_{\ell},s_{\ell}}).
\end{equation}
Calibration uses only source-scenario score statistics and raw scores within the target block.
Here $\mathcal{I}_{u,s}=\{(r,i):r\in\mathcal{R}_{u,s},1\leq i\leq n_{u,s,r}\}$ denotes the target turns in the user--scenario block $(u,s)$.
It does not change the memory contents or expose target gold labels.
We use two calibration variants.

\paragraph{Mean-shift calibration.}
Let $\mathcal{B}_{u,s}$ denote a user--target-scenario block of size $|\mathcal{B}_{u,s}|$, and let $\ell$ index a target turn in this block.
Define the block-level mean raw score as:
\begin{equation}
  \bar{\hat{y}}_{\mathcal{B}_{u,s}}
  =
  \frac{1}{|\mathcal{B}_{u,s}|}
  \sum_{\ell'\in\mathcal{B}_{u,s}}\hat{y}_{\ell'}.
\end{equation}
Mean-shift calibration adjusts the block so that its mean matches the user's historical mean $\mu^{\mathrm{hist}}_{u,-s}$,
\begin{equation}
  \tilde{y}_{\ell}^{\mathrm{MS}}
  =
  \operatorname{clip}_{1,5}\!\left(
  R\left(\hat{y}_{\ell}+\mu^{\mathrm{hist}}_{u,-s}-\bar{\hat{y}}_{\mathcal{B}_{u,s}}\right)\right).
\end{equation}
Here $R(\cdot)$ rounds a real-valued score to the nearest integer score.

\paragraph{CDF calibration.}
This variant preserves the within-block rank order of raw scores and maps each rank to the inverse empirical CDF of the user's source-scenario scores,
\begin{equation}
  \tilde{y}_{\ell}^{\mathrm{CDF}}
  =
  F^{-1}_{u,-s}\!\left(\frac{\operatorname{rank}_{\mathcal{B}_{u,s}}(\hat{y}_{\ell})+0.5}{|\mathcal{B}_{u,s}|}\right).
\end{equation}
Here $\operatorname{rank}_{\mathcal{B}_{u,s}}(\hat{y}_{\ell})$ is the zero-based rank of $\hat{y}_{\ell}$ among raw scores in the target block, and $F^{-1}_{u,-s}$ is the inverse empirical CDF of the user's source-scenario gold scores.

The main results use CDF-calibrated scores, while ablations compare raw, mean-shift, and CDF variants to separate judging quality from score-scale correction.

\section{Personalized Satisfaction Evaluator Verification}
\label{sec:evaluator-verification}


\subsection{Experimental Setup}
\label{sec:data-protocol}

\begin{table*}[!t]
\centering
\small
\begin{threeparttable}
\begin{tabular}{llcccc}
\toprule
\multirow{2}{*}{Type} & \multirow{2}{*}{Evaluator}
& \multicolumn{3}{c}{Score agreement}
& \multicolumn{1}{c}{DSAT detection} \\
\cmidrule(lr){3-5}
\cmidrule(lr){6-6}
& & Pearson $\uparrow$ & Spearman $\uparrow$ & QWK $\uparrow$ & F1 $\uparrow$ \\
\midrule
\multirow{2}{*}{Supervised scorer}
& BERT scorer & 0.0126 & 0.0192 & 0.0125 & 0.2268 \\
& BERT ordinal scorer & 0.0363 & 0.0625 & 0.0322 & 0.1984 \\
\midrule
\multirow{2}{*}{User-history retrieval}
& Nearest-history score & 0.3281 & 0.3529 & 0.2992 & 0.2361 \\
& RAG prompted scorer & 0.1360 & 0.0776 & 0.0282 & 0.0214 \\
\midrule
\multirow{5}{*}{Generic LLM judge}
& SPUR-style evaluator\tnote{a} & 0.1314 & 0.1095 & 0.0842 & 0.2955 \\
& Zero-shot judge & 0.1865 & 0.1274 & 0.1753 & 0.1732 \\
& Few-shot judge & 0.1924 & 0.1578 & 0.1867 & 0.2680 \\
& Task-rubric judge & 0.1600 & 0.1210 & 0.1511 & 0.2559 \\
& Prometheus-rubric judge & 0.2066 & 0.1435 & 0.2005 & 0.2207 \\
\midrule
\multirow{1}{*}{User-aware LLM judge}
& \textbf{User-memory evaluator} & \textbf{0.3601} & \textbf{0.3716} & \textbf{0.3595} & \textbf{0.3655}
\\
\bottomrule
\end{tabular}
\begin{tablenotes}[flushleft]
\footnotesize
\item[a] SPUR-style evaluation produces binary labels, which are mapped to scores 4 and 3 for ordinal metrics.
\end{tablenotes}
\caption{Meta-evaluation of satisfaction evaluators against human turn-level satisfaction annotations. Score-agreement metrics evaluate 1--5 satisfaction prediction, while F1 evaluates dissatisfied-turn detection. \textbf{Bold} indicates the best result in each column.}
\label{tab:main-baseline-comparison}
\end{threeparttable}
\vspace{-1em}
\end{table*}

\paragraph{Data and protocol.}
This section evaluates whether the proposed evaluator can approximate human turn-level satisfaction judgments under a personalized setting.
The main evaluation draws on the same real-user Chinese task-oriented conversation collection as \citet{10.1145/3767695.3769490}, using its user profiles, task scenarios, conversation trajectories, turn-level 1--5 satisfaction scores, and dissatisfaction reasons for low-satisfaction turns.
We use these satisfaction annotations as gold labels for evaluator verification.
Additional details of the collection protocol are provided in Appendix~\ref{sec:appendix-data-collection}.

The evaluation follows a cross-scenario protocol.
For a target scenario $s$, the evaluator may use the same user's labeled conversations from other scenarios, but not labels from the target scenario:
let $\mathcal{S}_{-s}=\mathcal{S}\setminus\{s\}$, and define
\begin{equation}
\mathcal{H}_{u}^{-s}=\{D_{u,s',r}:s'\in\mathcal{S}_{-s}, r\in\mathcal{R}_{u,s'}\}.
\end{equation}
This setting tests whether user-specific preferences and scoring styles transfer across task scenarios without leaking target-scenario labels.
Each target example is an assistant-turn example $\ell=(u,s,r,i)$ and is evaluated against the human satisfaction score $y_{\ell}$.
The evaluation split contains 90 users, 356 user--scenario blocks, and 6,474 assistant turns from four planning-oriented task families: recipe planning, gift preparation, travel planning, and skill learning planning.

We also use URS, A User-Centric Multi-Intent Benchmark for Evaluating Large Language Models~\citep{wang-etal-2024-user}, as a supplementary validation resource.
Because URS provides session-level satisfaction rather than the same turn-level annotation structure as the main dataset, it is reported separately in Appendix~\ref{sec:appendix-urs-validation}.

\paragraph{Baselines.}
The baselines are chosen to test three alternatives to personalized semantic evaluation.
Supervised baselines learn score prediction from labeled data, represented by BERT and ordinal BERT models~\citep{devlin-etal-2019-bert}.
Retrieval-based baselines reuse the same user's labeled source-scenario turns as non-parametric personalized evidence.
Generic judging baselines use rubric-based or prompted LLM judges without personalized evidence, including SPUR-style rubric induction~\citep{lin2024interpretable} and zero-shot, few-shot, task-rubric, and Prometheus-style LLM-as-a-judge variants~\citep{liu2023g,zheng2023judging,kim2024prometheus}.
All LLM-backed methods in the main experiments use Qwen3-8B~\citep{yang2025qwen3} as the backbone, and generic LLM judges do not receive user profiles, source-scenario histories, or user memory.
In contrast to generic judges, our evaluator uses structured user memory built from the source-scenario history before scoring the target turn.
Implementation details for the baselines and evaluator variants are provided in Appendix~\ref{sec:appendix-baseline-details}.

\paragraph{Metrics.}
We report complementary measures of agreement with human satisfaction judgments.
Pearson correlation measures linear agreement on the 1--5 scale, Spearman correlation measures rank agreement, and quadratic weighted kappa evaluates ordinal agreement with stronger penalties for larger score differences.
We also report F1-DSAT for the dissatisfied class induced by the threshold between scores 3 and 4, since missing dissatisfied turns is particularly harmful for personalized conversation evaluation.
Additional metric definitions are provided in Appendix~\ref{sec:appendix-metrics}.

\subsection{Evaluator Performance}

Table~\ref{tab:main-baseline-comparison} shows that the memory-based evaluator obtains the strongest agreement with human satisfaction judgments across all reported metrics.
Compared with the best generic LLM-as-a-judge baseline, the memory evaluator substantially improves both ordinal agreement and dissatisfied-turn detection.
This result indicates that judging the local response quality alone is not sufficient for personalized satisfaction evaluation, even when the same backbone is used.

Retrieval-based methods are stronger than generic judges on ranking metrics, which confirms that labeled user history contains useful evidence about individual scoring behavior.
However, the strongest retrieval baseline remains much weaker on F1-DSAT than the memory evaluator.
This gap suggests that reusing nearby historical labels can capture score-scale similarity, but it does not replace a semantic evaluator that conditions on both user memory and the current assistant response.

The supervised BERT baselines perform poorly on ordinal agreement.
This behavior is consistent with the user-specific nature of the task: the labeled data are limited relative to the user-by-scenario diversity that a supervised model would need to fit.
The result supports the use of a training-free evaluator that represents user history explicitly rather than relying only on a compact supervised encoder.

\subsection{Additional Analysis}

We further examine how user memory, score calibration, and backbone capacity affect evaluator reliability.
Table~\ref{tab:compact-evaluator-ablation} reports a compact ablation on the 20-user subset, with the complete variant table provided in Appendix~\ref{sec:appendix-evaluator-ablation}.

\begin{table}[!t]
\centering
\footnotesize
\setlength{\tabcolsep}{3pt}
\begin{tabular}{llrrr}
\toprule
Model & Variant & Pearson & QWK & DSAT F1 \\
\midrule
\multirow{4}{*}{Qwen3-8B}
& No mem. & 0.139 & 0.071 & 0.046 \\
& Mem. & 0.300 & 0.281 & 0.351 \\
& Mem. + MS & 0.353 & 0.339 & 0.393 \\
& Mem. + CDF & 0.338 & 0.337 & 0.405 \\
\midrule
Qwen3.6 & Mem. + CDF & 0.398 & 0.397 & 0.412 \\
GPT-5.4m & Mem. + CDF & \textbf{0.421} & \textbf{0.420} & \textbf{0.416} \\
\bottomrule
\end{tabular}
\caption{Ablation analysis on the 20-user subset. Qwen3.6 denotes Qwen3.6-35B-A3B, and GPT-5.4m denotes GPT-5.4-mini. Mem., MS, and CDF denote structured user memory, mean-shift calibration, and CDF calibration.}
\label{tab:compact-evaluator-ablation}
\end{table}

\paragraph{User memory drives the largest qualitative change in evaluator behavior.}
Without memory, the judge has little agreement with human ratings and almost fails to identify dissatisfied turns.
Adding structured memory substantially improves both ordinal agreement and DSAT detection, suggesting that the evaluator benefits from representing how a specific user has previously assigned satisfaction scores rather than only judging the local response.

\paragraph{Calibration is complementary to memory, not a substitute.}
Mean-shift calibration is more effective for correcting the overall score scale, while CDF calibration better preserves the rank structure of turns within a user--scenario block and yields the strongest dissatisfied-turn detection among the Qwen3-8B variants.
This is why the main comparison in Table~\ref{tab:main-baseline-comparison} uses the CDF-calibrated evaluator.

\paragraph{Backbone capacity also matters.}
When the same memory and CDF calibration setting is applied to stronger judge backbones, the agreement metrics generally improve over Qwen3-8B.
These results indicate that stronger LLM backbones may provide more reliable satisfaction estimates, although the benchmark results in this paper are reported with Qwen3-8B for cost-effective and reproducible evaluation.

\FloatBarrier

\section{PersTurnBench: Personalized Turn-Level User Conversation Satisfaction Benchmark}
\label{sec:persturnbench}

We use the verified conversation satisfaction evaluator to instantiate PersTurnBench, a controlled benchmark for personalized turn-level user conversation satisfaction evaluation.
PersTurnBench asks how a candidate generation model would affect user satisfaction when it is placed into the same historical conversation state.
Unlike standard response-quality benchmarks, the evaluation state includes user-specific evidence and the score is interpreted as predicted satisfaction for the current user turn.
Because candidate responses are evaluated without being rolled into later turns, different generation models are compared under the same conversation prefixes, user histories, and evaluator state for each replay item.

\subsection{Benchmark Protocol}

\begin{figure*}[t]
\centering
\includegraphics[width=0.88\linewidth]{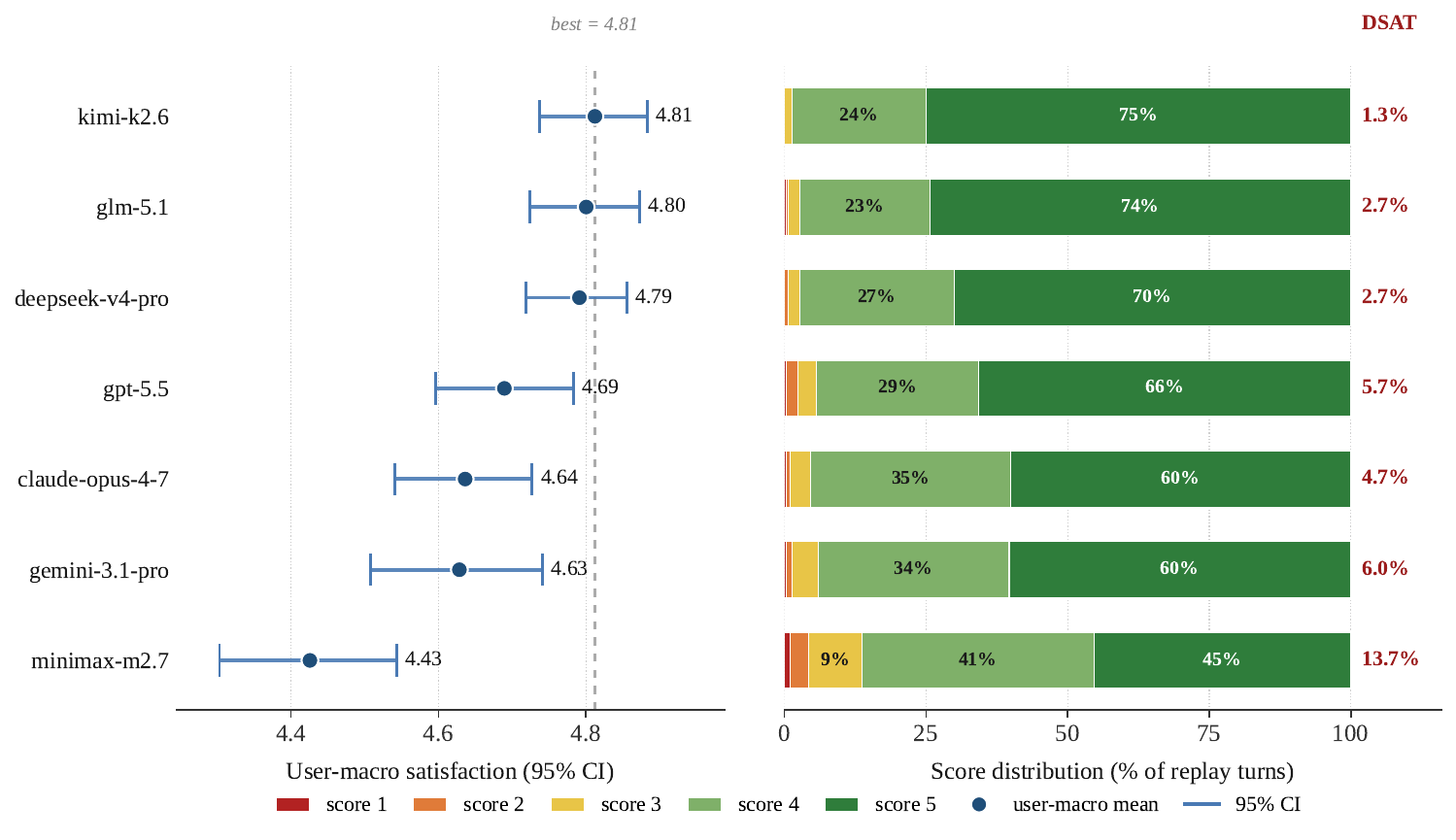}
\caption{PersTurnBench results under reference-CDF calibrated scoring. The left panel reports user-macro satisfaction with 95\% confidence intervals, and the right panel shows the calibrated 1--5 score distribution. Red labels denote predicted dissatisfied-turn rates, defined as the fraction of turns with scores at most 3.}
\label{fig:persturnbench-main}
\vspace{-1em}
\end{figure*}

Each benchmark item is a turn-level counterfactual evaluation instance.
It contains a user, a target scenario, a historical conversation prefix, the current user request, the task context, and the source-scenario user history available to the evaluator.
The candidate model fills only the current assistant response slot.
This design keeps the user state fixed and isolates the effect of replacing the assistant response on the predicted satisfaction.

We evaluate all candidate models on the same replay subset, which contains 300 turns from 90 users and 180 user--task blocks.
In the main benchmark track, candidate models generate responses given the conversation prefix.
The frozen evaluator is the Qwen3-8B memory evaluator introduced in Section~\ref{sec:evaluator}.
For every candidate response, we apply the same evaluator and aggregate the resulting satisfaction estimates across replay items.

We use the reference-CDF calibrated evaluator score as the official PersTurnBench reporting score.
This calibrated view maps candidate scores through user-specific reference score distributions, reducing raw score inflation while preserving model-level separation.
Figure~\ref{fig:persturnbench-main} reports the main benchmark result with user-macro satisfaction and dissatisfied-turn rate, while Appendix~\ref{sec:appendix-persturnbench-additional} provides the full aggregate table and additional comparisons.

The primary metric is user macro satisfaction, which first averages predicted satisfaction within each user and then averages across users.
This avoids letting users with more replay turns dominate the benchmark score.
We additionally report micro mean satisfaction, task macro mean, user--task block macro mean, SAT rate, and DSAT rate in the appendix.

\subsection{Benchmark Results}

Figure~\ref{fig:persturnbench-main} shows that PersTurnBench separates candidate models into several broad groups rather than producing a single uniformly spaced ranking.
\texttt{kimi-k2.6}, \texttt{glm-5.1}, and \texttt{deepseek-v4-pro} form the strongest group, with high user-macro satisfaction and low dissatisfied-turn rates.
\texttt{gpt-5.5}, \texttt{claude-opus-4-7}, and \texttt{gemini-3.1-pro} form a middle group whose confidence intervals overlap more substantially.
\texttt{minimax-m2.7} is separated most clearly from the other models, especially by its higher DSAT rate.

The comparison between user-macro satisfaction and DSAT rate illustrates why PersTurnBench reports both average satisfaction and dissatisfaction risk.
Average scores capture the overall quality of generated responses under personalized states, while DSAT rate highlights cases where a model is predicted to produce explicitly unsatisfactory turns.
This distinction is important for personalized conversation systems because a small change in average score may still correspond to a noticeable difference in the frequency of user-level failures.

The confidence intervals in Figure~\ref{fig:persturnbench-main} are also part of the benchmark signal.
They are wider than a turn-level-only aggregation would suggest because PersTurnBench aggregates over users before averaging.
This reflects the intended evaluation target: a model should perform well not only across replay turns but also across different users whose satisfaction criteria may vary.

Appendix~\ref{sec:appendix-persturnbench-additional} provides the complete aggregate metric table, pairwise comparisons, and memory-augmented candidate generation results under the same replay setting.
The pairwise comparisons support the aggregate view by showing whether a model's gains are broad or concentrated in a smaller subset of turns.
The memory-augmented results further illustrate how PersTurnBench can evaluate candidate systems that expose additional user-history context to the generator while keeping the evaluator and replay states fixed.

\section{Conclusion}


We studied personalized turn-level user conversation satisfaction evaluation, which estimates how a specific user would rate a target assistant response under prior history and current context.
We developed a conversation satisfaction evaluator that summarizes past ratings and interaction histories into user memory, applies this memory during turn-level judging, and calibrates scores to match user-specific rating scales.
Meta-evaluation against human satisfaction annotations shows that the evaluator improves over supervised, retrieval-based, and generic LLM-as-a-judge baselines, especially when memory and post-hoc calibration are combined.
We further introduced PersTurnBench, a replay-based benchmark that reuses the verified evaluator as a frozen judge to compare candidate generation models under fixed personalized conversation states.
PersTurnBench provides a reproducible, lower-cost screening layer before user studies, but it is not a replacement for direct human evaluation.
These results show that personalized satisfaction evaluators expose turn-level differences and user-level disagreement that generic judges miss.
Future work can extend this framework to multilingual settings, interactive multi-turn evaluation, and integration with personalization systems.

\section{Limitations}


This work has several limitations that delimit the intended scope of the evaluator and benchmark.

First, the main conversations are in Chinese and the participant pool skews toward student users.
The observed satisfaction patterns may therefore reflect Chinese-language conventions and the preferences of this particular pool, and may not fully transfer to other populations or multilingual settings.
The evaluated scenarios are deliberately scoped to planning-oriented tasks, where shared structure across users supports controlled comparison; extending the framework to broader assistant use cases is a natural next step.

Second, the conversation satisfaction evaluator is an automatic judge and therefore an approximation of direct human satisfaction.
Although user memory and calibration substantially improve agreement with human annotations over generic LLM-judge and retrieval baselines, absolute human-level agreement on personalized turn-level satisfaction remains an open challenge, and PersTurnBench should be viewed as a reproducible screening layer that complements rather than substitutes direct user studies.
The replay protocol further focuses on single-turn replacements under fixed conversation states, which is intentionally designed for controlled cross-model comparison and leaves multi-turn user adaptation to interactive evaluation studies.
Post-hoc calibration is based on the assumption that historical rating distributions provide a useful reference scale for future satisfaction estimates, which we expect to hold most strongly within the same task family.
These limitations point to broader user populations, languages, domains, and interactive evaluation protocols as the main directions for future extensions.

\section{Ethical Considerations}


This work uses conversation data and satisfaction annotations collected from real users.
The collection protocol was reviewed within our laboratory before data collection began, with attention to participant consent, compensation, and anonymization procedures.
Participants were recruited through public recruitment posts and social media channels.
They were informed that their conversations and feedback could be used for research and analysis, and participation in the study indicated consent to the collection.
Participants were compensated per completed conversation at an average rate of approximately 8--10 RMB.

The collection process required contact and payment information, such as names, phone numbers, and bank account information, for compensation.
These identifiers are not included in the data used in this paper.
The paper uses anonymized user identifiers, task contexts, conversation trajectories, structured user profiles, and satisfaction annotations.
The profile fields used by the study are based on tag selection or multiple-choice questions rather than open-ended personal descriptions.

The main risk of this work is misuse of an automatic satisfaction evaluator as a substitute for direct user feedback.
We therefore frame PersTurnBench as a reproducible screening benchmark rather than a replacement for human evaluation or deployment-time user studies.
Because the data are collected in Chinese and the participant pool skews toward student users, the evaluator may also reflect population-specific preferences and should be validated before being used in other languages, communities, or high-stakes settings.
The artifacts introduced in this paper are intended for research on personalized conversation evaluation and should not be used as a deployment-time substitute for direct consented user feedback.
External models, datasets, and software artifacts are used only for research evaluation and implementation under their respective access terms.
AI assistance was used during paper writing, language polishing, and experimental development, and all resulting content, code, and analyses were checked by the authors.

\bibliography{custom}

\appendix
\clearpage
\section{Notation Summary}
\label{sec:appendix-notation}


Table~\ref{tab:notation-summary} summarizes the main notation used for task formulation, evaluator construction, calibration, and benchmark evaluation.

\begin{table}[h]
\centering
\small
\setlength{\tabcolsep}{4pt}
\renewcommand{\arraystretch}{1.08}
\begin{tabular}{@{}p{0.25\columnwidth}p{0.67\columnwidth}@{}}
\toprule
Symbol & Meaning \\
\midrule
$\mathcal{U}$ & Set of users. \\
$\mathcal{S}$ & Set of task scenarios. \\
$u,s,r,i$ & User, scenario, session, and turn indices. \\
$p_u$ & Profile of user $u$. \\
$\mathcal{R}_{u,s}$ & Conversation sessions collected from user $u$ under scenario $s$. \\
$\tau_{s,r}$ & Concrete task context for session $r$ under scenario $s$. \\
$D_{u,s,r}$ & Conversation session for user $u$, scenario $s$, and session $r$. \\
$\mathcal{H}_{u}^{-s}$ & Source-scenario conversation history available for user $u$ before evaluating scenario $s$. \\
$\ell=(u,s,r,i)$ & Target assistant-turn example. \\
$C_{\ell}$ & Conversation prefix before the target assistant response. \\
$q_{\ell}$ & Current user request at target example $\ell$. \\
$a_{\ell}$ & Assistant response being evaluated at target example $\ell$. \\
$y_{\ell}$ & Human satisfaction score in $\{1,2,3,4,5\}$. \\
$z_{\ell}$ & Binary satisfaction label, where $z_{\ell}=1$ iff $y_{\ell}\geq 4$. \\
$\mathcal{E}$ & Personalized turn-level satisfaction evaluator. \\
$\hat{y}_{\ell}$ & Raw predicted satisfaction score. \\
$\hat{z}_{\ell}$ & Predicted binary satisfaction label induced from $\hat{y}_{\ell}$. \\
$\tilde{y}_{\ell}$ & Post-hoc calibrated satisfaction score. \\
$m_u^{-s}$ & User memory built from $\mathcal{H}_{u}^{-s}$ and $p_u$. \\
$g_{\theta}$ & Memory builder. \\
$\mathcal{J}_{\phi}$ & LLM judge used by the evaluator. \\
$d_{\ell}$ & Natural-language diagnostic rationale. \\
$\kappa$ & Post-hoc score calibration function. \\
$\mathcal{B}_{u,s}$ & User--target-scenario block for calibration and benchmark aggregation. \\
$F^{-1}_{u,-s}$ & Inverse empirical CDF of user $u$'s source-scenario gold scores. \\
\bottomrule
\end{tabular}
\caption{Notation used in the evaluator and benchmark formulation.}
\label{tab:notation-summary}
\end{table}

\section{Main Data Collection Protocol}
\label{sec:appendix-data-collection}


\paragraph{Participants and recruitment.}
The main evaluation data come from the real-user Chinese task-oriented assistant conversations also used by \citet{10.1145/3767695.3769490}.
We recruited 115 participants through public recruitment posts and social media channels.
Participants were native Chinese speakers with prior experience using LLM-based assistants.
The recruitment material briefly described the study as an online experiment on LLM conversation satisfaction, stated that participants would complete personalized conversation tasks and evaluate assistant responses, and explained that the data could be used for research and analysis.
By registering for the study and completing the online tasks, participants indicated consent to this data collection and research use.
Participants were compensated per completed conversation at an average rate of approximately 8--10 RMB.

\paragraph{Profile and task setup.}
Each participant first completed a structured user profile.
The profile included gender, age, occupation, personality, daily interests, travel habits, dining preferences, spending habits, and other preference tags.
Except for age, the profile fields were collected through tag selection or multiple-choice questions rather than open-ended personal descriptions.
The collection covered four planning-oriented task families: recipe planning, gift preparation, travel planning, and skill learning planning.
Each participant could complete up to 19 tasks across the four scenarios, with roughly four to five tasks per scenario on average.

\begin{table}[t]
\centering
\scriptsize
\begin{tabular}{ll}
\toprule
Statistic & Value \\
\midrule
Participants & 115 \\
Task families & 4 \\
Conversations & 1,833 \\
User turns & 8,204 \\
Assistant turns with satisfaction labels & 8,060 \\
Average conversations per participant & 15.94 \\
Average assistant turns per conversation & 4.40 \\
Average satisfaction score & 4.23 \\
Satisfied / dissatisfied-or-neutral turns & 6,799 / 1,261 \\
\midrule
Recipe planning conversations & 402 \\
Gift preparation conversations & 531 \\
Travel planning conversations & 502 \\
Skill learning conversations & 398 \\
\bottomrule
\end{tabular}
\caption{Statistics of the main conversation data used in this paper.
Satisfied turns are defined as turns with satisfaction scores of 4 or 5, and dissatisfied-or-neutral turns are those with scores of 1--3.}
\label{tab:main-data-statistics}
\end{table}

Table~\ref{tab:main-data-statistics} summarizes the resulting data scale.

\paragraph{Conversation instructions.}
For each task, the interface presented a concrete task background before the conversation started.
Participants were instructed to read the task background carefully, follow their own real needs and preferences, and complete at least the requirements specified by the task.
They were also encouraged to add additional requirements when appropriate and to ask follow-up questions about details that they cared about.
The goal was for the final plan or recommendation to be usable in a realistic situation.
Participants could end a conversation when they believed the task had been completed or when continuing the conversation would no longer be helpful, but the instructions encouraged substantive multi-turn interaction rather than ending after only one or two turns.
No strict minimum or maximum number of turns was imposed.

\paragraph{Annotation instructions.}
After each conversation, participants completed a questionnaire for that conversation.
For each assistant turn, they annotated hallucination and satisfaction.
The satisfaction label used a 1--5 scale with the following anchors: 1 means very dissatisfied and not helpful, 2 means dissatisfied and not sufficient for decision making, 3 means neutral with some inspiration but insufficient detail, 4 means satisfied and helpful but still improvable, and 5 means very satisfied with no obvious better response.
When the satisfaction score was at most 3, participants selected the primary dissatisfaction reason from five predefined options: insufficient detail, insufficient diversity, failure to satisfy the requirement, unusable in practice, and other.
The dissatisfaction reason was therefore a categorical choice rather than free text.
The questionnaire also included conversation-level diagnostic questions about memory, detail, availability, and diversity.

\paragraph{Quality and privacy handling.}
We did not discard conversations only because they were difficult or yielded low satisfaction scores.
Conversations that were extremely short or were judged to be invalid submissions (e.g., empty or nonsensical content) were excluded from the paid set, but conversations were not penalized for content judgments such as low satisfaction scores or task difficulty.
Payment-related identifiers such as names, phone numbers, and bank account information were stored separately from the research data and are not included in the paper experiments.
The experimental version used in this paper keeps anonymous user identifiers, task contexts, structured profiles, conversation trajectories, and satisfaction annotations.
We also ran an automatic pattern scan for common direct identifiers such as phone numbers, bank-card-like numbers, identity-card-like numbers, and email addresses.
The matches we found were public or generated contact information inside assistant responses, such as hotel or restaurant contact details, rather than participant identity or payment information.
This scan does not guarantee that no personal information appears in free-form conversation text, but it reduces the risk of direct participant identifiers being present in the version used for experiments.

\section{Additional Metrics}
\label{sec:appendix-metrics}


For full 1--5 satisfaction evaluation, we additionally compute absolute error metrics.
Given gold scores $y_i$ and predictions $\hat{y}_i$ over $N$ target turns, mean absolute error and root mean squared error are
\begin{align}
  \mathrm{MAE} & = \frac{1}{N}\sum_{i=1}^{N}|y_i-\hat{y}_i|, \\
  \mathrm{RMSE} & = \sqrt{\frac{1}{N}\sum_{i=1}^{N}(y_i-\hat{y}_i)^2}.
\end{align}
Pearson correlation measures linear score association:
\begin{equation}
  r =
  \frac{\sum_i (y_i-\bar{y})(\hat{y}_i-\bar{\hat{y}})}
       {\sqrt{\sum_i (y_i-\bar{y})^2}\sqrt{\sum_i(\hat{y}_i-\bar{\hat{y}})^2}}.
\end{equation}
Quadratic weighted kappa measures ordinal agreement while penalizing larger score disagreements more heavily:
\begin{equation}
  \kappa_{\mathrm{QW}} =
  1 - \frac{\sum_{a,b} w_{ab} O_{ab}}{\sum_{a,b} w_{ab} E_{ab}},
  w_{ab}=\frac{(a-b)^2}{(K-1)^2},
\end{equation}
where $O$ is the observed confusion matrix, $E$ is the expected confusion matrix under independent marginals, and $K=5$.
We also report false-SAT and false-DSAT rates for the 3/4 satisfaction boundary:
\begin{align}
  \mathrm{FalseSAT} & =
  \frac{\#\{i: y_i\leq 3,\ \hat{y}_i\geq 4\}}
       {\#\{i: y_i\leq 3\}}, \\
  \mathrm{FalseDSAT} & =
  \frac{\#\{i: y_i\geq 4,\ \hat{y}_i\leq 3\}}
       {\#\{i: y_i\geq 4\}}.
\end{align}
For personalized analysis, within-user centered metrics first remove each user's mean score:
\begin{equation}
  y_i^{c}=y_i-\bar{y}_{u(i)},
  \hat{y}_i^{c}=\hat{y}_i-\bar{\hat{y}}_{u(i)}.
\end{equation}

\section{Baseline and Evaluator Details}
\label{sec:appendix-baseline-details}


This section summarizes the implementation details of the baseline families and the main evaluator used in Section~\ref{sec:evaluator-verification}.
All methods are evaluated on the same target turns and use the same gold satisfaction annotations for meta-evaluation.

\paragraph{Supervised baselines.}
The BERT baseline trains a supervised encoder on the available training split and predicts the 1--5 satisfaction label for each target assistant turn.
The ordinal BERT variant uses the same input information but changes the output formulation to better respect the ordered score scale.
These baselines do not construct explicit user memory and therefore mainly test whether a compact supervised model can learn the evaluation task from the available labeled data.

\paragraph{History and retrieval baselines.}
Distributional baselines estimate target scores from train-set or user-history score statistics, including mean, median, majority, and empirical score-distribution variants.
Retrieval baselines search the same user's source-scenario turns and use nearest labeled examples as non-parametric personalized evidence.
The raw nearest-retrieval variant directly uses retrieved historical labels, while the RAG prompt variant provides retrieved examples to an LLM judge and asks it to produce an absolute satisfaction score.

\paragraph{Generic LLM-as-a-judge baselines.}
Generic judging baselines score the target assistant response from the target context only.
They do not receive the user profile, source-scenario history, or constructed user memory.
We include zero-shot, few-shot, task-rubric, and Prometheus-style rubric variants to test whether stronger generic criteria can replace personalized evidence.

\paragraph{SPUR-style baseline.}
The SPUR-style baseline induces a generic scoring rubric and applies it to target assistant turns without user-specific memory.
Because this baseline is primarily boundary-oriented, we map its SAT/DSAT output to the nearest 1--5 scores for the full-score table and interpret it mainly as a dissatisfaction-boundary baseline.

\paragraph{Main evaluator.}
The main conversation satisfaction evaluator uses Qwen3-8B with structured user memory and no target-scenario memory update.
For each user--target-scenario block, the evaluator first builds memory from the same user's labeled source-scenario histories.
It then scores each target assistant turn with the fixed memory, target conversation context, task context, and assistant response.
Raw evaluator outputs are reported together with mean-shift and CDF-calibrated variants when applicable.

\subsection{Implementation Details}
\label{sec:appendix-implementation-details}

All main prompt-based evaluator experiments use Qwen3-8B~\citep{yang2025qwen3} as the LLM backbone and are run through an OpenAI-compatible API interface.
For local runs, the model is served with vLLM~\citep{kwon2023efficient}, and the main low-cost validation setting uses a maximum model context length of 16,384 tokens.
The same backbone is used for memory construction and turn-level satisfaction evaluation unless a backbone ablation explicitly changes one of them.
The memory is built once for each user--target-scenario block from source-scenario histories and is then kept fixed while evaluating the target block.
The target-turn evaluator uses a recent-context window of five previous conversation messages.
LLM structured-output calls use a low-temperature setting of 0.3 for the main memory-based evaluator, while generic LLM-as-a-judge baselines use temperature 0.2.
We did not conduct a broad hyperparameter search for prompt-based evaluators; prompt variants, memory variants, and calibration variants are treated as controlled ablations.

For post-hoc calibration, mean-shift calibration aligns the predicted block mean with the historical user--scenario reference mean before rounding to the 1--5 scale.
Reference-CDF calibration maps each predicted score through the empirical distribution of predictions on the corresponding reference block and then maps the percentile to the empirical gold-score distribution of that block.
The benchmark results in the main text use the reference-CDF calibrated score as the reported evaluator score.

The supervised BERT baselines use \texttt{bert-base-chinese}~\citep{devlin-etal-2019-bert}.
Unless otherwise stated, they use a maximum input length of 512 tokens, at most 12 conversation turns in the serialized context, batch size 16, evaluation batch size 32, five epochs, learning rate $2\times10^{-5}$, weight decay 0.01, warmup ratio 0.06, dropout 0.1, and random seed 42.
The ordinal variant uses the same encoder and input fields as the classification variant, with an ordinal decision threshold of 0.5.
Checkpoints are selected by validation MAE.

For PersTurnBench candidate generation, candidate LLMs receive the fixed replay prefix under the benchmark context condition.
Candidate responses are generated with temperature 0.7 and a maximum generation length of 1,024 tokens in the main collection script.
For memory-augmented candidate-generation analyses, the retrieval modes use four retrieved cross-scenario conversation memories by default, with a maximum of 700 assistant-response characters per retrieved item and four previous local messages stored in each memory record.
Replay scoring uses the same frozen evaluator configuration across all candidate models so that model comparisons differ only in the candidate response being evaluated.

\paragraph{Computing infrastructure.}
All local experiments were run on NVIDIA A100-SXM4-80GB GPUs.
We estimate that the reported local experiments required approximately 200 A100 GPU-hours in total, excluding inference performed through external API endpoints.
This estimate includes evaluator verification, supervised baseline training, local candidate generation, replay scoring, calibration analyses, and appendix ablations.

\paragraph{Packages and parameters.}
The supervised baselines are implemented with PyTorch~\citep{Ansel_PyTorch_2_Faster_2024} and Hugging Face Transformers~\citep{wolf-etal-2020-transformers}.
Dataset splits use group-based user splits implemented with scikit-learn~\citep{scikit-learn}.
Correlation metrics use SciPy~\citep{2020SciPy-NMeth}, while F1, kappa, and error metrics use scikit-learn.
TF-IDF retrieval baselines and memory-augmented generation contexts use scikit-learn's \texttt{TfidfVectorizer} and cosine similarity with default vectorizer settings unless otherwise stated.
All OpenAI-compatible model calls use the same prompt templates reported in Appendix~\ref{sec:appendix-prompts}.
We provide code and scripts at \url{https://github.com/wzf2000/PersTurnBench}.

\section{Evaluator Ablations}
\label{sec:appendix-evaluator-ablation}


This section expands the compact ablation analysis in Table~\ref{tab:compact-evaluator-ablation}.
The appendix table reports all available mean-shift and CDF calibration variants for each analyzed backbone.
All reported rows are evaluated on the same 20-user personalized test subset, containing 1,594 turn-level examples.
The no-memory row is included for Qwen3-8B to isolate the contribution of structured user memory under the main open-source backbone.

\begin{table*}[t]
\centering
\small
\begin{threeparttable}
\begin{tabular}{llcccc}
\toprule
Backbone & Evaluator variant & Pearson $\uparrow$ & Spearman $\uparrow$ & QWK $\uparrow$ & F1-DSAT $\uparrow$ \\
\midrule
\multirow{4}{*}{Qwen3-8B}
& No memory & 0.1392 & 0.1069 & 0.0706 & 0.0459 \\
& Structured memory & 0.2999 & 0.3012 & 0.2807 & 0.3505 \\
& Structured memory + MS & \textbf{0.3529} & 0.3737 & \textbf{0.3386} & 0.3934 \\
& Structured memory + CDF & 0.3384 & \textbf{0.3760} & 0.3374 & \textbf{0.4050} \\
\midrule
\multirow{3}{*}{Qwen3.6-35B-A3B}
& Structured memory & 0.3411 & 0.3335 & 0.3083 & 0.4168 \\
& Structured memory + MS & 0.3828 & 0.4028 & 0.3758 & \textbf{0.4207} \\
& Structured memory + CDF & \textbf{0.3977} & \textbf{0.4238} & \textbf{0.3966} & 0.4121 \\
\midrule
\multirow{3}{*}{GPT-5.4-mini}
& Structured memory & 0.3706 & 0.3319 & 0.3393 & 0.3987 \\
& Structured memory + MS & 0.4093 & 0.4014 & 0.3967 & 0.3993 \\
& Structured memory + CDF & \textbf{0.4212} & \textbf{0.4199} & \textbf{0.4200} & \textbf{0.4156} \\
\bottomrule
\end{tabular}
\caption{Full ablation table for evaluator design choices and backbone models on the same 20-user personalized test subset.}
\label{tab:evaluator-ablation}
\end{threeparttable}
\end{table*}

\section{Supplementary URS Validation}
\label{sec:appendix-urs-validation}


This section reports supplementary validation on URS~\citep{wang-etal-2024-user}.
Unlike the main turn-level setting, URS provides user-reported satisfaction at the session level rather than turn-level satisfaction annotations.
We therefore adapt the evaluation unit from individual assistant turns to complete sessions and treat the results as a cross-resource sanity check rather than a direct counterpart to the main turn-level meta-evaluation.

\begin{table*}[t]
\centering
\small
\begin{threeparttable}
\begin{tabular}{llrrrrr}
\toprule
Category & Method & MAE $\downarrow$ & Pearson $\uparrow$ & Spearman $\uparrow$ & QWK $\uparrow$ & F1-DSAT $\uparrow$ \\
\midrule
\multirow{2}{*}{No memory}
& Direct prompt & 1.0497 & 0.1588 & 0.1295 & 0.1396 & 0.5152 \\
& Calibrated prompt & 0.7928 & 0.1621 & 0.1527 & 0.1411 & 0.3415 \\
\midrule
\multirow{2}{*}{Memory baseline}
& Structured memory prompt & 0.8305 & 0.2312 & 0.2354 & 0.1868 & \textbf{0.5386} \\
& Calibrated memory prompt & 0.7260 & 0.2358 & 0.2495 & 0.2073 & 0.4592 \\
\midrule
Ours
& Task-guarded evaluator & \textbf{0.6952} & \textbf{0.2760} & \textbf{0.2883} & \textbf{0.2424} & 0.5207 \\
\bottomrule
\end{tabular}
\begin{tablenotes}[flushleft]
\footnotesize
\item All rows use the URS test split with 584 session-level examples.
\end{tablenotes}
\caption{Supplementary URS session-level validation results.}
\label{tab:urs-validation}
\end{threeparttable}
\end{table*}

The task-guarded evaluator is the strongest URS result among the tested variants.
Compared with the calibrated memory baseline, it improves MAE from 0.7260 to 0.6952, Pearson from 0.2358 to 0.2760, QWK from 0.2073 to 0.2424, and F1-DSAT from 0.4592 to 0.5207.

These results support the same qualitative message as the main experiments: personalized context and task-aware scoring can improve automatic satisfaction estimation.
At the same time, URS remains harder than the main turn-level setting because it lacks explicit turn-level labels and user profiles, and because session-level satisfaction compresses multiple interaction effects into one score.

\section{Additional PersTurnBench Replay Results}
\label{sec:appendix-persturnbench-additional}


\subsection{Aggregate Replay Scores}

Table~\ref{tab:persturnbench-refcdf} reports the full aggregate metric table corresponding to Figure~\ref{fig:persturnbench-main}.
All scores use reference-CDF calibrated evaluator outputs.

\begin{table*}[t]
\centering
\scriptsize
\setlength{\tabcolsep}{4pt}
\begin{threeparttable}
\begin{tabular}{lrrrrrr}
\toprule
Candidate model
& Micro $\uparrow$
& User macro (95\% CI) $\uparrow$
& Task macro $\uparrow$
& Block macro $\uparrow$
& SAT rate $\uparrow$
& DSAT rate $\downarrow$ \\
\midrule
\texttt{kimi-k2.6} & 4.737 & \textbf{4.812 [4.737, 4.883]} & 4.731 & 4.785 & 0.987 & 0.013 \\
\texttt{glm-5.1} & 4.707 & 4.800 [4.724, 4.873] & 4.697 & 4.760 & 0.973 & 0.027 \\
\texttt{deepseek-v4-pro} & 4.667 & 4.791 [4.718, 4.855] & 4.655 & 4.732 & 0.973 & 0.027 \\
\texttt{gpt-5.5} & 4.573 & 4.689 [4.596, 4.783] & 4.564 & 4.624 & 0.943 & 0.057 \\
\texttt{claude-opus-4-7} & 4.540 & 4.636 [4.541, 4.727] & 4.524 & 4.597 & 0.953 & 0.047 \\
\texttt{gemini-3.1-pro} & 4.527 & 4.628 [4.508, 4.741] & 4.532 & 4.583 & 0.940 & 0.060 \\
\texttt{minimax-m2.7} & 4.263 & 4.426 [4.303, 4.543] & 4.252 & 4.344 & 0.863 & 0.137 \\
\bottomrule
\end{tabular}
\caption{Detailed PersTurnBench aggregate results with reference-CDF calibrated scores.}
\label{tab:persturnbench-refcdf}
\end{threeparttable}
\end{table*}

\subsection{Pairwise Replay Comparisons}

\begin{figure}[t]
    \centering
    \includegraphics[width=1\linewidth]{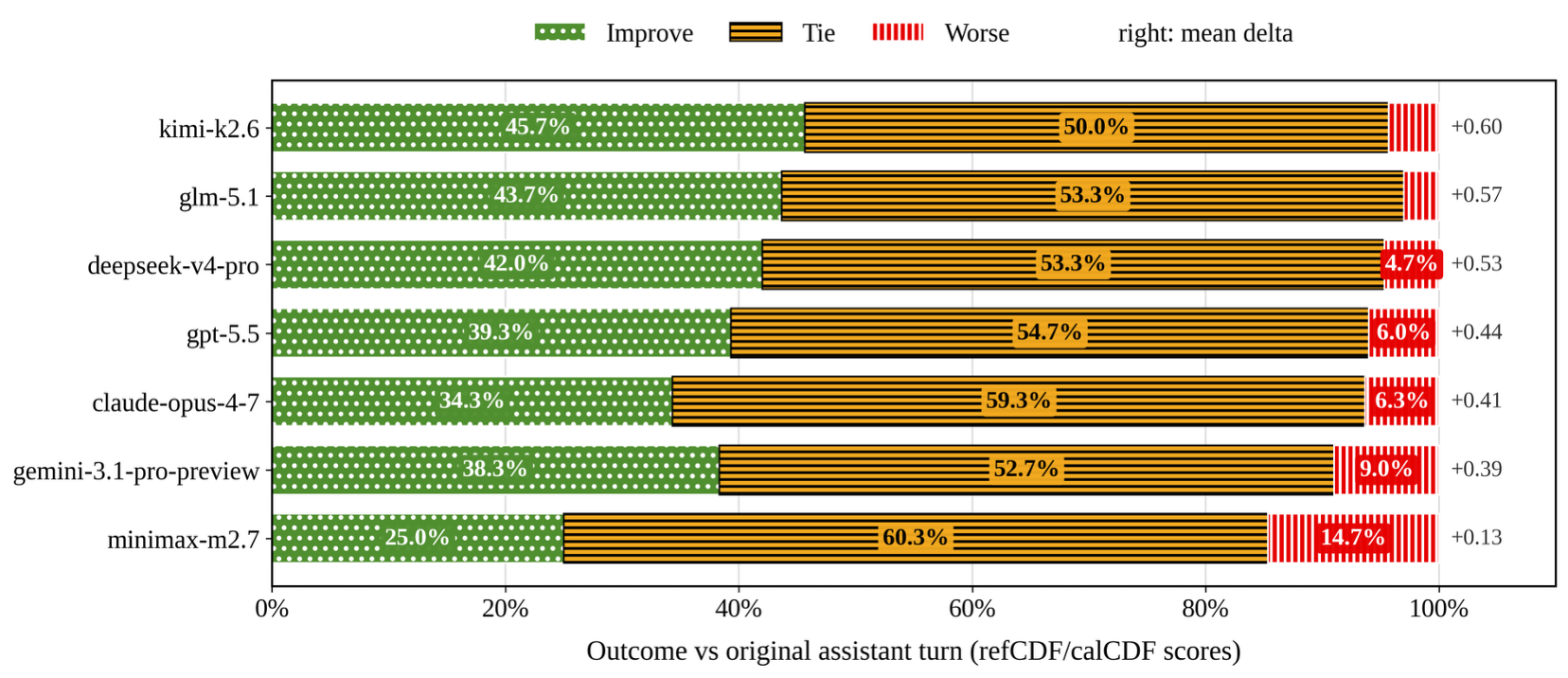}
    \caption{Pairwise comparison with original responses from the conversation data.}
    \label{fig:pair-wise-original}
\end{figure}

\begin{figure}[t]
    \centering
    \includegraphics[width=1\linewidth]{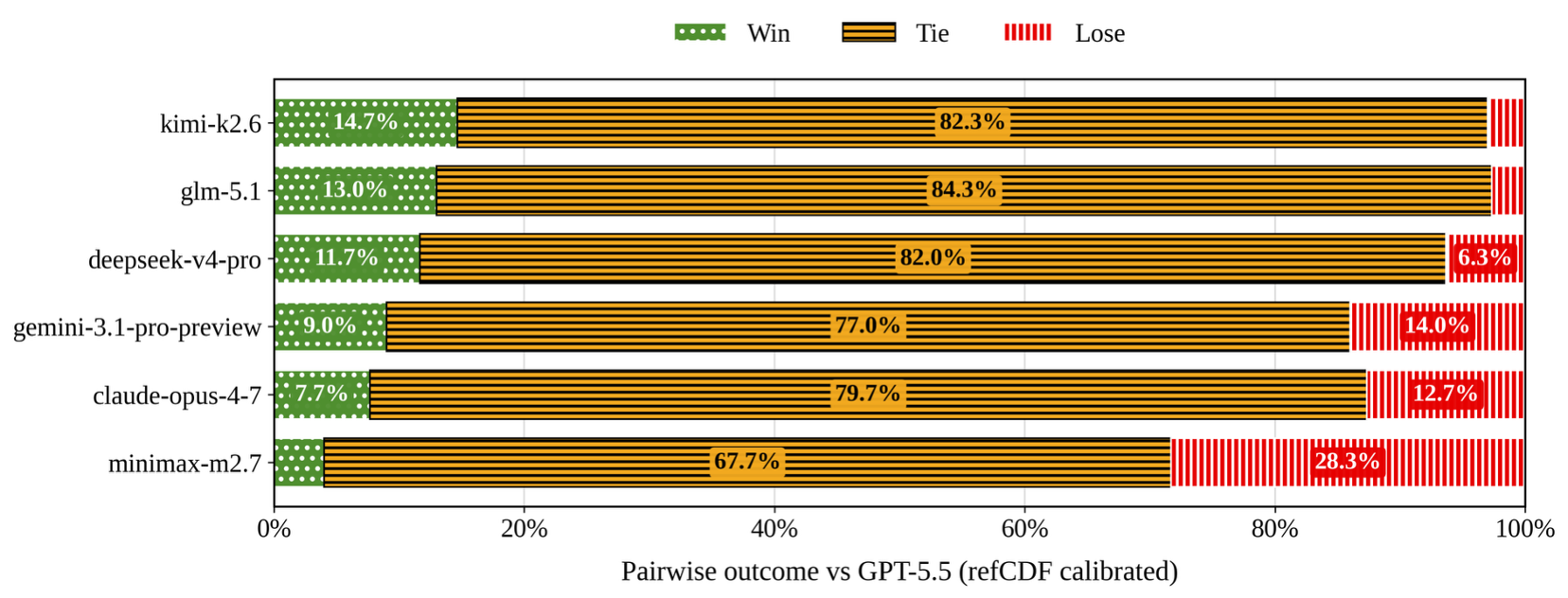}
    \caption{Pairwise comparison with \texttt{gpt-5.5} in PersTurnBench.}
    \label{fig:pair-wise-gpt-5.5}
\end{figure}

This subsection reports pairwise replay outcomes that complement the aggregate PersTurnBench scores in Figure~\ref{fig:persturnbench-main} and Table~\ref{tab:persturnbench-refcdf}.
Figure~\ref{fig:pair-wise-original} compares each candidate model against the original assistant responses in the conversation data.
Figure~\ref{fig:pair-wise-gpt-5.5} compares other candidate models against \texttt{gpt-5.5} under the same reference-CDF calibrated scoring view.

Compared with the original assistant turns, the top models improve roughly 42--46\% of replay cases while losing only a small fraction of cases.
\texttt{minimax-m2.7} has a much lower improvement rate and a higher worse rate, which is consistent with its lower mean replay scores in Table~\ref{tab:persturnbench-refcdf}.
When using \texttt{gpt-5.5} as the pairwise reference, most models have large tie regions, but \texttt{kimi-k2.6}, \texttt{glm-5.1}, and \texttt{deepseek-v4-pro} still show more wins than losses.
This pairwise view is useful because it exposes whether a higher aggregate score comes from broad small improvements or from a smaller number of large changes.

\subsection{Memory-Augmented Candidate Generation}

We also evaluate whether exposing retrieved user conversation memories to candidate generation models improves PersTurnBench scores.
The candidate-visible memory contains raw source-scenario conversation snippets from the same user, but does not include satisfaction scores, dissatisfaction reasons, or user profile annotations.
We compare prefix-only generation with two memory-augmented modes: TF-IDF retrieval and diverse retrieval.
All rows use the same 300-turn replay subset, the same frozen Qwen3-8B memory evaluator, and reference-CDF calibrated scores.

\begin{table*}[t]
\centering
\scriptsize
\setlength{\tabcolsep}{5pt}
\begin{tabular}{llrrrr}
\toprule
Candidate model & Generation context & Micro $\uparrow$ & User macro $\uparrow$ & SAT rate $\uparrow$ & DSAT rate $\downarrow$ \\
\midrule
\multirow{3}{*}{\texttt{Qwen3-8B}}
& Prefix only & \textbf{4.333} & \textbf{4.465} & \textbf{0.920} & \textbf{0.080} \\
& TF-IDF memory & 4.273 & 4.371 & 0.903 & 0.097 \\
& Diverse memory & 4.260 & 4.383 & 0.897 & 0.103 \\
\midrule
\multirow{3}{*}{\texttt{deepseek-v4-flash}}
& Prefix only & \textbf{4.320} & \textbf{4.421} & 0.893 & 0.107 \\
& TF-IDF memory & 4.250 & 4.415 & 0.883 & 0.117 \\
& Diverse memory & 4.290 & 4.414 & \textbf{0.900} & \textbf{0.100} \\
\midrule
\multirow{3}{*}{\texttt{gemini-3.1-flash-lite}}
& Prefix only & \textbf{4.203} & \textbf{4.339} & \textbf{0.870} & \textbf{0.130} \\
& TF-IDF memory & 4.177 & 4.290 & 0.867 & 0.133 \\
& Diverse memory & 4.167 & 4.310 & \textbf{0.870} & \textbf{0.130} \\
\bottomrule
\end{tabular}
\caption{PersTurnBench results for prefix-only and memory-augmented candidate generation contexts.}
\label{tab:persturnbench-memory-generation}
\end{table*}

Table~\ref{tab:persturnbench-memory-generation} shows that naive memory augmentation does not improve the current benchmark scores.
Across the three tested candidate models, prefix-only generation has the highest micro and user macro scores.
The drop is small but consistent for Qwen3-8B and Gemini, while diverse retrieval is closer to prefix-only than TF-IDF retrieval for DeepSeek.
This suggests that simply prepending retrieved raw conversation memories is not sufficient for better personalized generation.
Although this negative result does not support naive memory prepending, it provides a useful baseline for future work: stronger memory-augmented systems should use selective or compressed user memory rather than long retrieved conversation snippets.

\section{Prompt Templates}
\label{sec:appendix-prompts}


This appendix reports the main prompt templates used for the primary dataset experiments.
The prompts are shown as templates with placeholders such as \texttt{\{user profile\}}, \texttt{\{history conversations\}}, and \texttt{\{assistant response\}}.
We include both the original Chinese prompt and an English translation for readability.

\subsection{Memory Construction}

\begin{promptbox}{Memory Construction Prompt: Chinese Original}
你是一名用户行为分析专家。
请基于以下用户的历史对话记录，建立一份精准的个性化用户记忆，用于预测该用户对未来助手回复的满意度。

【用户画像】\{user profile\}

【满意度统计】总轮数：\{number of turns\}，平均分：\{average score\}，分布：1分$\times$\{n1\} / 2分$\times$\{n2\} / 3分$\times$\{n3\} / 4分$\times$\{n4\} / 5分$\times$\{n5\}

--- 历史对话（按任务顺序）---

\{history conversations with turn-level satisfaction scores and low-score reasons\}

--- 按分数分组的对比证据（重点参考）---

\{examples grouped by scores 5, 4, 3, 2, and 1\}

--- 分析任务 ---

请严格基于以上对比证据完成以下分析，不得使用对所有用户都成立的泛化描述：

1. 【评分边界 4$\rightarrow$5】：对比 5 分和 4 分轮次，指出哪些具体要素决定了能否从 4 分升至 5 分（必须引用上面的实际例子）。
2. 【评分边界 3$\rightarrow$4】：对比 4 分和 3 分（及以下）轮次，指出导致从 4 分跌至 3 分的具体缺陷类型。
3. 【评分风格】：该用户是偏严格还是偏宽松？结合平均分给出校准说明。
4. 【用户特异性要求】：该用户有哪些一般用户没有的特定要求？如果所有用户都会这样要求，则不算特异性。
5. 【偏好格式】：该用户偏好什么回复结构或组织形式？
6. 【任务观察】：各任务类型下有哪些特殊偏好？

可参考的不满意原因类别：其它、不够多样、不可用、不够细致、不满足需求。

请严格按照 JSON Schema 输出，不要输出其他内容。
\end{promptbox}

\begin{promptbox}{Memory Construction Prompt: English Translation}
You are a user behavior analyst.
Based on the following historical conversations of a user, build an accurate personalized user memory for predicting this user's future satisfaction with assistant responses.

[User profile] \{user profile\}

[Satisfaction statistics] Number of turns: \{number of turns\}; average score: \{average score\}; distribution: score 1 $\times$ \{n1\} / score 2 $\times$ \{n2\} / score 3 $\times$ \{n3\} / score 4 $\times$ \{n4\} / score 5 $\times$ \{n5\}.

=== Historical conversations in task order ===

\{history conversations with turn-level satisfaction scores and low-score reasons\}

=== Comparative evidence grouped by score ===

\{examples grouped by scores 5, 4, 3, 2, and 1\}

=== Analysis task ===

Complete the following analysis strictly based on the comparative evidence above.
Do not use generic descriptions that would apply to all users.

1. [Boundary from 4 to 5]: Compare score-5 and score-4 turns, and identify the concrete factors that determine whether a response can move from 4 to 5.
You must cite actual examples above.
2. [Boundary from 3 to 4]: Compare score-4 and score-3-or-lower turns, and identify the concrete defect types that make a response fall from 4 to 3.
3. [Scoring style]: Is this user strict or lenient?
Use the average score to calibrate the conclusion.
4. [User-specific requirements]: What requirements are specific to this user and not merely generic expectations shared by all users?
5. [Preferred format]: What response structure or organization does this user prefer?
6. [Task observations]: What task-specific preferences appear in different task types?

Candidate dissatisfaction reasons: other, insufficient diversity, unusable, insufficient detail, failure to satisfy the requirement.

Output strictly according to the JSON schema and do not output anything else.
\end{promptbox}

\subsection{Memory-Based Turn Evaluation}

\begin{promptbox}{Memory-Based Turn Evaluation Prompt: Chinese Original}
你是一名个性化满意度评分员。
任务是给当前助手回复打 1--5 分。
请严格按下面 checklist 判断，不要写长篇分析。

【用户评分摘要】

评分风格：\{scoring style\}

历史平均分：\{average score\}

4分基线：\{three-vs-four boundary\}

5分门槛：\{four-vs-five boundary\}

用户特定要求：

\{user-specific requirements\}

偏好回复形式：\{preferred response format\}

任务特定观察：

\{task-specific observations\}

【用户画像】\{user profile\}

【任务背景】\{task context\}

【最近对话历史】

\{recent conversation history\}

【待评估的助手回复】

\{assistant response\}

【可选原因标签】其它、不够多样、不可用、不够细致、不满足需求、满意。

【只按这 3 步判断】

Step 1. 先判断是否达到 4 分基线。
若没有直接回答问题、明显忽略约束、帮助性不足，给 1/2/3。

Step 2. 若已达到 4 分，再判断是否满足 5 分门槛。
只有明显满足关键细节、格式和用户特定要求时才给 5。
只要整体合格但还缺少关键一项，就给 4。

Step 3. 选择一个最贴切的原因标签。

请严格输出 JSON，不要输出其他内容：

\{
  "classification": 1--5 中的整数,
  "reason": "若 classification $\geq$ 4 则为满意；否则为一个不满意原因",
  "analysis": "用 2--4 句写明是否过 4 分基线、是否满足 5 分门槛以及最终分数依据"
\}
\end{promptbox}

\begin{promptbox}{Memory-Based Turn Evaluation Prompt: English Translation}
You are a personalized satisfaction scorer.
Your task is to assign a 1--5 score to the current assistant response.
Follow the checklist below strictly and do not write a long analysis.

[User scoring summary]

Scoring style: \{scoring style\}

Historical average score: \{average score\}

Score-4 baseline: \{three-vs-four boundary\}

Score-5 threshold: \{four-vs-five boundary\}

User-specific requirements:

\{user-specific requirements\}

Preferred response format: \{preferred response format\}

Task-specific observations:

\{task-specific observations\}

[User profile] \{user profile\}

[Task context] \{task context\}

[Recent conversation history]

\{recent conversation history\}

[Assistant response to evaluate]

\{assistant response\}

[Candidate reason labels] other, insufficient diversity, unusable, insufficient detail, failure to satisfy the requirement, satisfied.

[Judge only with the following three steps]

Step 1. First decide whether the response reaches the score-4 baseline.
If it does not directly answer the question, clearly ignores constraints, or is insufficiently helpful, assign 1/2/3.

Step 2. If the response reaches the score-4 baseline, decide whether it satisfies the score-5 threshold.
Assign 5 only when it clearly satisfies the key details, format, and user-specific requirements.
If the response is generally acceptable but misses one key item, assign 4.

Step 3. Select the most appropriate reason label.

Output strict JSON and do not output anything else:

\{
  "classification": an integer from 1 to 5,
  "reason": "satisfied if classification $\geq$ 4; otherwise one dissatisfaction reason",
  "analysis": "Use 2--4 sentences to explain whether the response passes the score-4 baseline, whether it satisfies the score-5 threshold, and why the final score is assigned."
\}
\end{promptbox}

\subsection{Generic LLM-as-a-Judge Baselines}

\begin{promptbox}{Generic LLM-as-a-Judge Prompt: Chinese Original}
系统消息：你是一名通用对话满意度评审器。
你只能根据当前任务、对话历史和待评分回复判断，不得使用任何用户画像或跨任务个性化记忆。

请预测用户对下面这轮助手回复的满意度，分数为 1--5。

评分标准：

1分：明显没有满足用户需求，严重不相关、不可执行、误导或遗漏关键约束。

2分：只满足少量需求，存在明显缺漏、泛泛而谈或实用性很弱。

3分：部分满足需求，但仍有较明显不足；处在不满意与满意边界的偏不满意侧。

4分：基本满足需求，结构清楚且可执行，仅有轻微不足；处在满意侧。

5分：高度满足或超出需求，具体、贴合、完整、可直接使用。

任务类型：\{task type\}

任务背景：\{task context\}

对话历史（待评分回复之前）：

\{conversation prefix\}

待评分助手回复：

\{assistant response\}

注意：ground truth 中 dissatisfied reason 只在 score $\leq$ 3 时生效；score $\geq$ 4 一律视为满意。

只输出 JSON 对象，不要输出其他文字：

\{
  "score": 1到5的整数,
  "reason": "若 score $\leq$ 3，从[其它, 不够多样, 不可用, 不够细致, 不满足需求]中选一个；若 score $\geq$ 4，必须为满意",
  "analysis": "一句话说明评分依据"
\}
\end{promptbox}

\begin{promptbox}{Generic LLM-as-a-Judge Prompt: English Translation}
System message: You are a generic conversation satisfaction judge.
You may judge only from the current task, conversation history, and response to be scored.
Do not use any user profile or cross-task personalized memory.

Predict the user's satisfaction with the following assistant response on a 1--5 scale.

Scoring rubric:

1: The response clearly fails to satisfy the user need; it is severely irrelevant, not actionable, misleading, or misses key constraints.

2: The response satisfies only a small part of the need and has clear omissions, generic content, or weak practical value.

3: The response partially satisfies the need but still has obvious shortcomings; it is on the dissatisfied side of the satisfied/dissatisfied boundary.

4: The response basically satisfies the need, is clearly structured and actionable, and has only minor shortcomings; it is on the satisfied side.

5: The response highly satisfies or exceeds the need; it is specific, well matched, complete, and directly usable.

Task type: \{task type\}

Task context: \{task context\}

Conversation history before the scored response:

\{conversation prefix\}

Assistant response to score:

\{assistant response\}

Note: in the ground truth, dissatisfied reasons apply only when score $\leq$ 3.
Scores $\geq$ 4 are always treated as satisfied.

Output only a JSON object and no other text:

\{
  "score": an integer from 1 to 5,
  "reason": "if score $\leq$ 3, choose one from [other, insufficient diversity, unusable, insufficient detail, failure to satisfy the requirement]; if score $\geq$ 4, it must be satisfied",
  "analysis": "one sentence explaining the scoring rationale"
\}
\end{promptbox}

\subsection{Episodic RAG Baseline}

\begin{promptbox}{Episodic RAG Prompt: Chinese Original}
你正在预测一个具体用户对当前助手回复的满意度。

核心设定：

- 输出 1--5 分整数，1 最差，5 最好。
- 3/4 是最重要的满意/不满意边界：1--3 表示不满意，4--5 表示满意。
- 4/5 是满意程度细分：只有当回复明显超过用户历史中 4 分样例时才给 5。
- 历史证据不是总结，而是该用户过往每个已标注 assistant turn 的原始记忆检索结果；请优先比较当前回复与这些相似历史样例的质量差异。
- 不要因为检索证据里高分或低分更多就机械跟随比例；重点判断当前回复相对证据的具体优劣。

【用户画像】

\{user profile\}

【当前任务背景】

\{task context\}

【当前对话上下文】

\{conversation prefix\}

【待评分助手回复】

\{assistant response\}

【检索到的该用户历史原始记忆证据】

\{retrieved labeled historical turns\}

请按以下步骤判断，但最终只输出 JSON：

1. 先判断当前回复是否跨过 3/4 满意边界。
2. 如果未跨过边界，在 1/2/3 中选择，并给出不满意原因。
3. 如果跨过边界，再比较是否只是合格满意 4，还是明显优于历史 4 分样例、可给 5。
4. reason 必须遵守：classification $\geq$ 4 时 reason 必须是“满意”；classification $\leq$ 3 时 reason 必须是不满意原因。

输出严格 JSON，不要 Markdown，不要额外文字：

\{
  "classification": 4,
  "reason": "满意",
  "analysis": "用1--3句话说明最关键证据和边界判断。",
  "boundary\_side": "sat",
  "evidence\_confidence": "medium"
\}
\end{promptbox}

\begin{promptbox}{Episodic RAG Prompt: English Translation}
You are predicting a specific user's satisfaction with the current assistant response.

Core setting:

- Output an integer score from 1 to 5, where 1 is worst and 5 is best.
- The 3/4 boundary is the most important satisfied/dissatisfied boundary: 1--3 means dissatisfied, and 4--5 means satisfied.
- The 4/5 boundary separates levels of satisfaction: assign 5 only when the response clearly exceeds the user's historical score-4 examples.
- The historical evidence is not a summary; it consists of retrieved raw memory records for the user's previously labeled assistant turns.
Prioritize comparing the current response with these similar historical examples.
- Do not mechanically follow the proportion of high-score or low-score retrieved evidence.
Focus on the concrete strengths and weaknesses of the current response relative to the evidence.

[User profile]

\{user profile\}

[Current task context]

\{task context\}

[Current conversation context]

\{conversation prefix\}

[Assistant response to score]

\{assistant response\}

[Retrieved raw historical memory evidence for this user]

\{retrieved labeled historical turns\}

Judge with the following steps, but output only JSON:

1. First decide whether the current response crosses the 3/4 satisfaction boundary.
2. If it does not cross the boundary, choose among 1/2/3 and provide a dissatisfaction reason.
3. If it crosses the boundary, compare whether it is merely qualified satisfaction with score 4 or clearly better than historical score-4 examples and thus deserves 5.
4. The reason must obey the following rule: if classification $\geq$ 4, reason must be "satisfied"; if classification $\leq$ 3, reason must be a dissatisfaction reason.

Output strict JSON, no Markdown and no extra text:

\{
  "classification": 4,
  "reason": "satisfied",
  "analysis": "Use 1--3 sentences to explain the key evidence and boundary judgment.",
  "boundary\_side": "sat",
  "evidence\_confidence": "medium"
\}
\end{promptbox}

\subsection{SPUR-Style Rubric Baseline}

\begin{promptbox}{SPUR-Style Rubric Scoring Prompt: Chinese Original}
系统消息：你是一名对话质量分析专家，擅长根据给定的评分标准判断用户满意度。
请仔细阅读满意/不满意的评判标准（rubric），再对给定对话进行综合判断。
只输出 JSON 对象，不输出其他内容。

\#\# 用户满意（SAT）的判断标准（rubric）：

\{induced SAT rubrics\}

\#\# 用户不满意（DSAT）的判断标准（rubric）：

\{induced DSAT rubrics\}

\#\# 待评估对话：

\{conversation\}

\#\# 任务：

请根据上述 rubric，对该对话进行逐条核对，并给出最终判断。

输出格式（严格遵守，不输出其他内容）：

\{
  "sat\_matches": [符合的SAT rubric编号列表，如 [1,3]],
  "dsat\_matches": [符合的DSAT rubric编号列表，如 [2]],
  "prediction": "SAT" 或 "DSAT",
  "confidence": 0.0--1.0,
  "reason": "简短的判断理由（一句话）"
\}
\end{promptbox}

\begin{promptbox}{SPUR-Style Rubric Scoring Prompt: English Translation}
System message: You are a conversation quality analyst who is good at judging user satisfaction according to given rubrics.
Read the satisfied and dissatisfied rubrics carefully, and then make an overall judgment for the given conversation.
Output only a JSON object and no other content.

\#\# Rubrics for user satisfaction (SAT):

\{induced SAT rubrics\}

\#\# Rubrics for user dissatisfaction (DSAT):

\{induced DSAT rubrics\}

\#\# Conversation to evaluate:

\{conversation\}

\#\# Task:

Check the conversation against the rubrics above one by one and give the final judgment.

Output format, strictly with no other content:

\{
  "sat\_matches": [indices of matched SAT rubrics, e.g., [1,3]],
  "dsat\_matches": [indices of matched DSAT rubrics, e.g., [2]],
  "prediction": "SAT" or "DSAT",
  "confidence": 0.0--1.0,
  "reason": "a short one-sentence rationale"
\}
\end{promptbox}

\begin{table*}[t]
\centering
\footnotesize
\setlength{\tabcolsep}{4pt}
\renewcommand{\arraystretch}{1.18}
\begin{tabularx}{\textwidth}{@{}>{\raggedright\arraybackslash}p{0.16\textwidth}>{\raggedright\arraybackslash}p{0.22\textwidth}>{\raggedright\arraybackslash}p{0.28\textwidth}>{\raggedright\arraybackslash}X@{}}
\toprule
\textbf{Case type} & \textbf{User annotation} & \textbf{Evaluator output} & \textbf{Observation} \\
\midrule
\textbf{Matches the core issue} & \textit{Score}: 3\newline \textit{Reason}: insufficient detail. & \textit{Score}: 3\newline \textit{Reason}: insufficient detail.\newline \textit{Rationale}: missing lodging, transportation, and timetable details. & The rationale points to the same missing actionable planning details selected by the user. \\
\addlinespace[3pt]
\textbf{Real issue but not the core reason} & \textit{Score}: 2\newline \textit{Reason}: failure to satisfy the requirement. & \textit{Score}: 3\newline \textit{Reason}: insufficient detail.\newline \textit{Rationale}: missing budget, price, and concrete itinerary details. & The rationale identifies a real weakness, but it misses that the user's primary issue is failure to satisfy the request. \\
\addlinespace[3pt]
\textbf{Incorrect or off-task rationale} & \textit{Score}: 2\newline \textit{Reason}: failure to satisfy the requirement. & \textit{Score}: 3\newline \textit{Reason}: failure to satisfy the requirement.\newline \textit{Rationale}: unmet preferences about travel and language-learning requirements. & The current turn is an anniversary-gift request, so the rationale imports requirements from unrelated source scenarios. \\
\bottomrule
\end{tabularx}
\caption{Qualitative rationale cases from cached evaluator outputs. Examples are shortened and translated to avoid exposing unnecessary user-specific details.}
\label{tab:rationale-case-study}
\end{table*}

\section{Rationale Case Study}
\label{sec:appendix-rationale-case-study}


The conversation satisfaction evaluator optionally outputs a short rationale together with the predicted satisfaction score.
The main experiments evaluate the score output against user satisfaction annotations, while the rationale is intended only as an auxiliary diagnostic signal.
This appendix provides a qualitative case study of these rationales rather than a full automatic evaluation, because the data contain user-selected dissatisfaction categories but do not contain free-form gold explanations.
Table~\ref{tab:rationale-case-study} summarizes representative cases.

We inspected evaluator outputs from the main Qwen3-8B memory evaluator on the personalized test split.
All evaluated turns contain a non-empty rationale, and the rationales usually identify concrete response-level issues such as missing details, missing budget information, weak structure, or failure to satisfy a user request.
Among low-satisfaction turns, the evaluator often assigns the correct broad dissatisfaction side, but its categorical reason predictions are biased toward insufficient detail.
This suggests that the rationale can be useful for understanding why the evaluator assigns a low score, but it should not be interpreted as a fully reliable explanation of the user's actual primary dissatisfaction reason.

We observed three recurring rationale patterns.
The first pattern shows rationales that match the user-selected core issue, for example when both annotation and evaluator point to missing actionable details.
A second, more common pattern identifies a real weakness in the response but misses the user's primary reason---for instance, explaining a ``failure to satisfy'' case as ``insufficient detail.''
A third pattern is off-task: the evaluator carries over evidence from source-scenario memory and mentions requirements from another scenario, which highlights that user memory improves score prediction but can also contaminate the generated explanation.

Overall, this case study supports treating the rationale as an interpretable byproduct of the evaluator rather than as a benchmark target.
The rationales can help diagnose model behavior and surface plausible failure modes, but future work should evaluate them with dedicated human judgments if they are used for explanation-facing applications.

\end{document}